\documentclass[lettersize,journal]{IEEEtran}
\usepackage{amsmath,amsfonts}
\usepackage{algorithmic}
\usepackage{algorithm}
\usepackage{array}
\usepackage[caption=false,font=small,labelfont=rm,textfont=rm]{subfig}

\usepackage{textcomp}
\usepackage{stfloats}
\usepackage{url}
\usepackage{verbatim}
\usepackage{graphicx}
\usepackage{cite}
\usepackage{subcaption}
\usepackage{cleveref}

\begin{document}

\title{CAMAv2: A Vision-Centric Approach for \\ Static Map Element Annotation}


\author{Shiyuan Chen$^{*}$, Jiaxin Zhang$^{*}$, Ruohong Mei, Yingfeng Cai, Haoran Yin, Tao Chen, Wei Sui and Cong Yang}%

\markboth{Journal of \LaTeX\ Class Files,~Vol.~14, No.~8, August~2021}%
{Shell \MakeLowercase{\textit{et al.}}: A Sample Article Using IEEEtran.cls for IEEE Journals}


%下面这一部分就是关键的首页图片的代码
\twocolumn[{
\renewcommand\twocolumn[1][]{#1}
\maketitle

\begin{center}   
    \centering
    \includegraphics[width=1.0\linewidth]{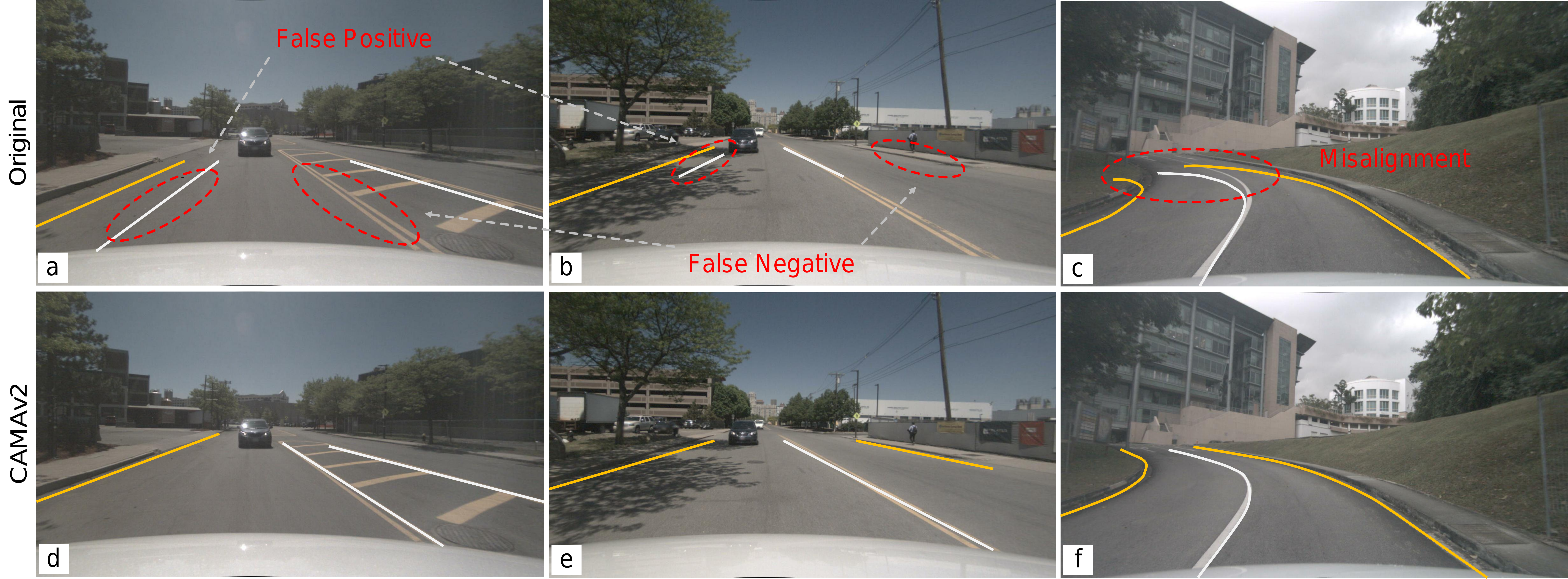}
    
    \captionof{figure}{\small Comparison of reprojection consistency and accuracy. The top and bottom lines present HD map reprojection of the original nuScenes (semi-automatic) and our CAMAv2 (pure-automatic) annotations. The yellow and white lines represent road boundaries and lane dividers, respectively. The original nuScenes has inconsistent road element annotations concerning the actual environments. For instance, there is no lane marking of the bikeway in images (a) and (b), which are wrongly indicated on the HD maps (a.k.a. False Negative). Besides, images (a) and (b) present lane divider and road boundary, but there is no HD map marking in the corresponding area (a.k.a. False Positive). Due to the lack of elevation information, the reprojected road elements are misaligned with the image (c). In contrast, the HD map from our proposed method shows better reprojection accuracy (f) and consistency (d, e). Best viewed in colour.
    }
   \label{fig:reprojection}
\end{center}%
}]
\let\thefootnote\relax\footnotetext{ S. Chen, J. Zhang, R. Mei, Y. Cai, H. Yin, T. Chen, W. Sui and C. Yang are with BeeLab, School of Future Science and Engineering, Soochow University, Suzhou, China. * Equal contribution. Corresponding: C. Yang, W. Sui and T. Chen (cong.yang@suda.edu.cn).}

% \maketitle

\begin{abstract}
The recent development of online static map element (a.k.a. HD map) construction algorithms has raised a vast demand for data with ground truth annotations. However, available public datasets currently cannot provide high-quality training data regarding consistency and accuracy. For instance, the manual labelled (low efficiency) nuScenes still contains misalignment and inconsistency between the HD maps and images (e.g., around 8.03 pixels reprojection error on average). To this end, we present CAMAv2: a vision-centric approach for Consistent and Accurate Map Annotation. Without LiDAR inputs, our proposed framework can still generate high-quality 3D annotations of static map elements. Specifically, the annotation can achieve high reprojection accuracy across all surrounding cameras and is spatial-temporal consistent across the whole sequence. We apply our proposed framework to the popular nuScenes dataset to provide efficient and highly accurate annotations. Compared with the original nuScenes static map element, our CAMAv2 annotations achieve lower reprojection errors (e.g., 4.96 vs. 8.03 pixels). Models trained with annotations from CAMAv2 also achieve lower reprojection errors (e.g., 5.62 vs. 8.43 pixels).

\end{abstract}


\section{Introduction}
\label{sec:intro}

\begin{figure*}[t!]
    \includegraphics[width=1\textwidth]{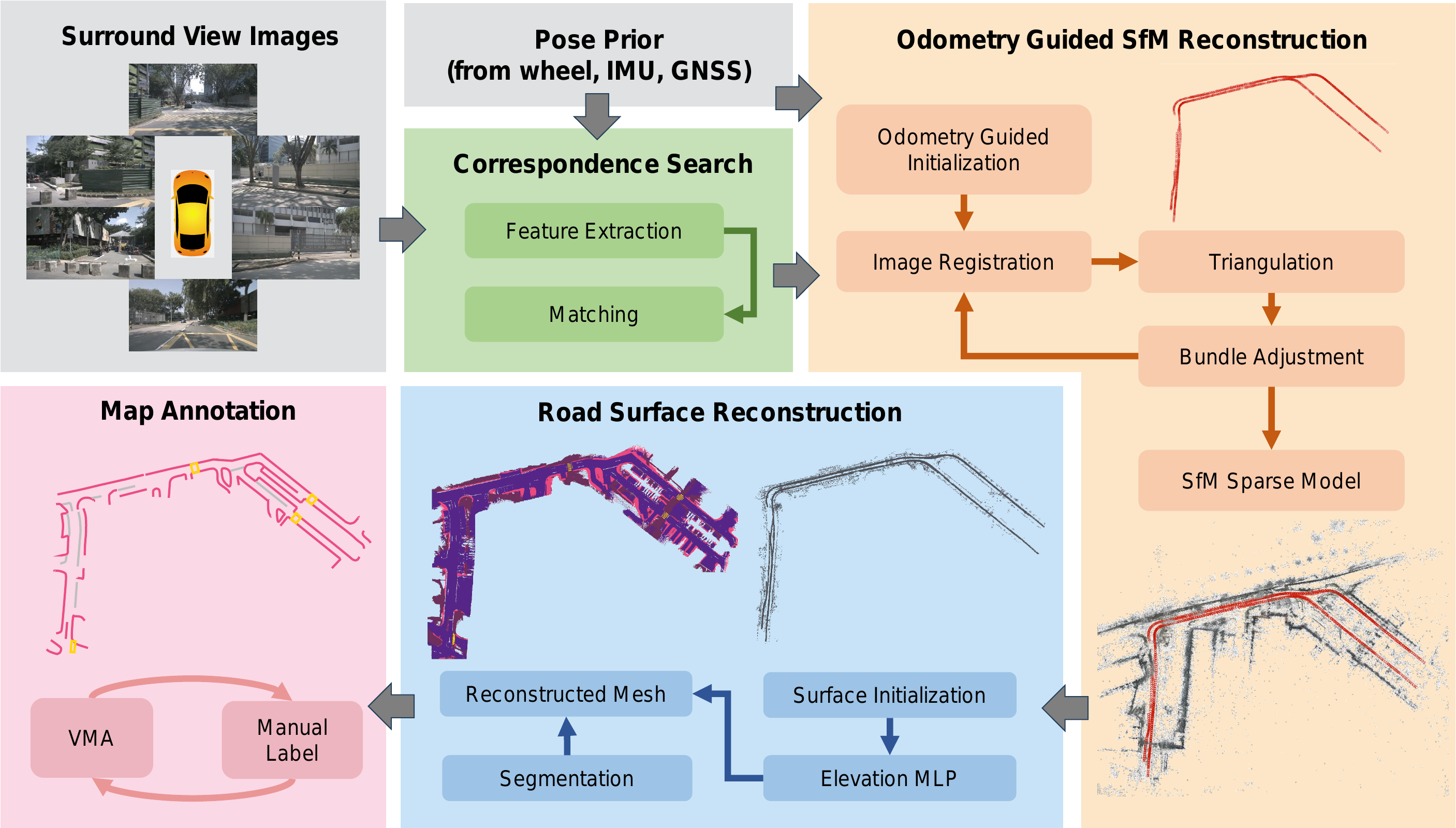}
  \caption{Illustration of our proposed reconstruction and annotation pipeline. The surround view images and auxiliary sensor data are fed into our proposed odometry-guided SfM to obtain highly accurate ego vehicle poses and sparse 3D points. A road surface mesh reconstruction called road surface reconstruction via mesh representations (RoMe) is applied to build dense 3D road surfaces with semantic labels. Finally, a vectorized map annotation (VMA) system is applied to produce a 3D HD map required by the perception algorithm as training data.}
  \label{fig:whole_pipeline}
  \vspace{-1em}
\end{figure*}

\IEEEPARstart{I}{ntelligent} vehicles use various sensors to detect surroundings, such as cameras and LiDAR. However, these sensors have a limited perception range and are vulnerable to bad weather. Using pre-built digital maps is an effective way to overcome the limitations, enhancing perception and robustness. High-definition (HD) maps provide a detailed and precise representation of the physical environment~\cite{tang2023high}, with lane-level instance and semantic information, which is vital for the navigation of intelligent vehicles. Most previous methods use simultaneous localisation and mapping (SLAM) for map construction, involving complicated pipelines and labour-intensive annotations~\cite{zhang2014loam, shan2018lego, shan2020lio}. The advancement of deep learning is driving a swift transformation in the technical architecture of intelligent driving perception algorithms, moving from traditional rule-based to data-driven methods~\cite{rao2018deep, gupta2021deep, zhou2023corner}. Online HD map construction is becoming the mainstream of LiDAR-based and vision-centric Bird's-Eye-View (BEV) perception~\cite{ma2022vision}. These methods focus on analysing vectorized static map element instances in BEV and 3D space, ultimately training neural networks to generate vectorized maps directly from surrounding camera images.

Existing online HD map construction algorithms usually require high-quality and diverse annotated training data. Accordingly, public datasets that provide annotations in 3D space can be roughly divided into two categories: HD map-based and depth reprojection-based. For instance, the nuScenes~\cite{caesar2020nuscenes} dataset provides human-annotated HD maps alongside ego vehicle poses. HDMapNet~\cite{li2022hdmapnet} is one of the pioneers that utilizes vectorized maps provided by nuScenes as ground truth and trains a neural network to predict static map elements directly in BEV space. OpenLane~\cite{chen2022persformer} proposed the first real-world 3D lane dataset using the depth reprojection method to generate 3D lane annotations. Specifically, each frame's 3D lane point clouds are generated by combining 2D lane detection and 3D LiDAR points.

However, there are still some significant limitations to the state-of-the-art annotation approaches. In particular, 3D road element annotations should accurately reflect real-world environments. To this end, we urgently need to address two key aspects to analyze existing annotations: consistency and geometric accuracy. The consistency focuses on the correspondence between the 3D annotations and the 2D images. The geometric accuracy reflects the matching accuracy in reprojecting the 3D annotations into the images. For example, Fig.~\ref{fig:reprojection} (a, b) shows the HD maps that plot the lane dividers between the vehicle and bike lanes. However, the images show no corresponding lane lines in the same area. Fig.~\ref{fig:reprojection} (c) shows the lane dividers projected on the image plane (white and yellow lines) deviating from the actual lane dividers in the image. The main reason is that the nuScenes dataset provides 2D HD maps without elevation information, and the ego-motion is not accurate when aligned with global maps. These limitations, therefore, affect the consistency and geometric accuracy of the 3D annotations with the actual 2D image.

% TODO rewrite
In light of these challenges, we propose \textbf{CAMAv2}: a vision-centric approach for \textbf{C}onsistent and \textbf{A}ccurate \textbf{M}ap \textbf{A}nnotation (see Fig.~\ref{fig:whole_pipeline}). Our proposed CAMAv2 is distinguished in three aspects: (1) We propose using a whole 3D reconstruction pipeline to get accurate camera motion and a sparse point cloud mainly from surround images. Thus, it can be applied to even low-cost intelligent driving platforms without equipping LiDAR. (2) A road surface mesh reconstruction algorithm~\cite{mei2023rome} is applied to reconstruct high-accuracy road surfaces. It can produce dense 3D road surfaces with both semantic and photometric information. (3) An auto map annotation tool~\cite{chen2023vma} is applied to extract the vectorized lane representation from the road surface. Consequently, as shown in Fig.~\ref{fig:reprojection} (bottom), CAMAv2 achieves high consistency and geometric accuracy compared with nuScenes default HD map.

The present article builds upon work first presented in~\cite{zhang2024cama}. Our contributions can be enumerated as follows.
\begin{itemize}
    \item We propose an efficient static element annotation framework, CAMAv2, for 3D road element annotation. The proposed CAMAv2 can generate highly consistent and geometric accurate HD map annotations. Through comprehensive experiments, we verified that such annotations can dramatically improve the accuracy and generalization of perception models in intelligent driving.
    \item To verify our proposed framework, we apply CAMAv2 to the nuScene dataset and set up new HD map annotations, namely nuScenes-CAMAv2. MapTRv2~\cite{liao2023maptrv2} is used as a benchmark model. Extensive experiments show that models trained on nuScenes-CAMAv2 can produce more consistent and accurate estimations of the static map elements compared with default HD map (e.g., reprojection errors of 5.62 vs. 8.43 pixels).
\end{itemize}

A preliminary version of this study was presented in~\cite{zhang2024cama}. The current journal extension introduces two major improvements. Firstly, we propose a parallel reconstruction method for large-scale intelligent driving scenarios, where multiple driving segments are first reconstructed separately and then stitched together, thus reducing the runtime required for SfM reconstruction (achieving five times efficiency boost). Second, we improve the reconstruction results of our pipeline on the nuScenes dataset by proposing a multi-scene aggregated reconstruction method, which solves the defects of dropping the head and tail frames in the previous single-scene reconstruction and the occlusion and blind zone problems. It ultimately results in more accurate and consistent annotations and greatly accelerates the annotation process. The nuScenes-CAMAv2 is publicly available at https://github.com/manymuch/CAMA.

\section{Related Work}
\label{sec:related}
Over the past few years, vision-centric BEV perception has become the 3D vision paradigm in intelligent driving. In particular, deep learning-based mapping methods have emerged as a prominent research interest in HD map construction. Effective view transformations are critical for improving algorithm performance and environment awareness. Data-driven BEV perception algorithms rely on high-accuracy and diverse 3D road surface element annotation. To the best of our knowledge, the public road surface element datasets provided static road elements can be roughly divided into HD map-based~\cite{caesar2020nuscenes, chang2019argoverse, wilson2023argoverse} and depth reprojection-based~\cite{chen2022persformer, yan2022once, li2023toponet}. In this part, we present a concise survey of existing HD map construction and correlated datasets. For a more thorough treatment of vision-centric HD map construction, a recent compilation by Ma {\it et al.}~\cite{ma2022vision} also offers sufficiently good reviews.

\subsection{Vision-centric HD Map Construction}
\label{s:related:hdmap}
HDMapNet~\cite{li2022hdmapnet} introduces a pioneering approach for online HD map construction by training neural networks to predict pixel-wise segmentation in BEV spaces, which requires complicated and time-consuming post-processing to obtain the vectorized representation of road elements. Following HDMapNet, VectorMapNet~\cite{liu2023vectormapnet} explores the end-to-end framework of HD map prediction, modeling map elements with a two-stage framework and predicting vectorized maps. MapTR~\cite{liao2022maptr, liao2023maptrv2} further boosts efficiency and performance by improving the map element decoder and loss function modeling. To make a step towards end-to-end road structure understanding, LaneGAP~\cite{liao2023lane} and TopoNet~\cite{li2023toponet} regress the road topology structure directly. In summary, the current trend of end-to-end mapping networks mainly focuses on improving the 2D-to-BEV transformation module and vectorized map element modeling method for greater effectiveness.
The view transformation between perspective view and BEV is usually conducted in geometric projection models explicitly~\cite{garnett20193d, philion2020lift, reiher2020sim2real} or uses neural networks to learn implicit representations~\cite{li2022bevformer, liu2022petr, chen2022efficient}. Data-driven perception algorithms drastically boost the advancement of intelligent driving applications. It offers several advantages. First, the model learns directly from the input through end-to-end training~\cite{diamond2021dirty}, eliminating the complex intermediate processing and enabling handle occlusion and extreme illuminance conditions through temporal fusion~\cite{huang2022bevdet4d, li2023bevstereo, huang2022bevpoolv2, wang2022mv}. Secondly, benefitting from the data-driven closed-loop mechanism, the models can self-optimize, reducing the engineering efforts in debugging corner cases. Thirdly, models trained on diverse datasets show better generalizability in various driving environments and complex visual conditions.% We refer the readers to this survey~\cite{ma2022vision} for a comprehensive development of the vision-centric HD map construction algorithm. 

\subsection{Map Element Datasets}
\label{s:related:dataset}
The nuScenes dataset~\cite{caesar2020nuscenes} provides four human-annotated city-scale maps, forming the basis for numerous online HD map construction works~\cite{li2022hdmapnet, liu2023vectormapnet, liao2022maptr}. All methods use three static map elements (lane boundary, lane divider, and pedestrian crossing) provided by the nuScenes HD map for training. Notably, nuScenes projects the reconstructed geometric map to the ground plane and annotates it in 2D space. Thus, the HD map lacks elevation information, and the reprojection accuracy between the HD map and images is not guaranteed. Fig.~\ref{fig:reprojection} shows the misalignment of the image with the reprojected road edges. In addition, due to the synchronization and calibration errors~\cite{zhang2021deep, zhang2022towards}, the pose between the cameras and the map is not well aligned. Consequently, the nuScenes HD map can not guarantee the consistency and accuracy of annotations.

The Argoverse2 dataset~\cite{wilson2023argoverse} is another large-scale dataset actively used for HD map construction. In contrast to the nuScenes dataset, Argoverse2 provides 3D HD map representation with high-resolution ground elevation and a map change dataset that depicts real-world HD map changes. Each scenario carries its local map region, and the advantages of per-scenario maps include more efficient queries and their ability to handle map changes. As it provides richer HD map information, more and more online mapping methods~\cite{liao2023maptrv2,ding2023pivotnet, qiao2023machmap} are evaluated on the Argoverse2 dataset.

Lanes are an important map element, and 3D lane detection has become a specialized perception task. Without HD maps, some datasets employ LiDAR points for 3D lane annotation, such as OpenLane~\cite{chen2022persformer} and Once-3DLanes~\cite{yan2022once}. They propose to combine LiDAR points and 2D lane segmentation to generate 3D lane annotation. The steps are as follows: Firstly, the corresponding relationship between 3D and 2D is established. The sparse depth information of 2D pixels is obtained by projecting the LiDAR points to the image plane. Then, the 2D lane segmentation results combined with sparse depth information are back-projected to 3D spaces and obtain lane point clouds. Finally, a filtering algorithm is applied to filtrate outliers and generate the 3D labelling results. This method guarantees geometric accuracy between 3D annotations and 2D images. However, annotating all the lanes in each frame does not impose spatial-temporal consistency in 3D space. Meanwhile, the LiDAR points could be noisy after multi-frame stitching due to the accumulation of localization errors and synchronization. Thus, the back-projected 3D lane may not be consistent with the real one.

% TODO rewrite
Unlike the above approaches, our proposed CAMAv2 reconstructs the static map element in a scene with surround view images (vision-centric). Without LiDAR and predefined HD maps, our approach can eliminate synchronization and calibration errors and impose spatial-temporal consistency in 3D space. Meanwhile, we propose a parallel reconstruction and spatial aggregation method for large-scale intelligent driving scenarios to speed up the reconstruction time. Our pipeline is verified on the nuScenes dataset, resulting in more accurate and consistent annotations.

\section{Approach}
\label{sec:approach}

Fig.~\ref{fig:whole_pipeline} illustrates the overview of our proposed CAMAv2. CAMAv2 is a vision-centric approach: the input is a set of surround view images, together with coarse ego poses obtained by auxiliary sensors (wheel, GNSS, and IMU). The whole framework mainly consists of two parts: Scene Reconstruction and Road Element Vector Annotation. 
The first part is fully automatic. To do so, we propose an improved Structure-from-Motion (SfM) that produces accurate sparse point clouds and ego poses for road surface initialization. Then, we apply road surface reconstruction via mesh representations (RoMe)~\cite{mei2023rome} to reconstruct the dense road surfaces. The second part is addressed semi-automatically based on a human-in-the-loop annotation. Mainly, the offline map auto-annotation model~\cite{chen2023vma} is first employed, followed by verification and modification by human annotators. Since CAMAv2 uses a reconstruction method based on image sequences that guarantee all 3D elements and their correspondence to 2D images, the 3D-2D correspondence and reprojection accuracy are also insured (even improved) without LiDAR.

\subsection{Scene Reconstruction}
As shown in Fig.~\ref{fig:whole_pipeline}, scene reconstruction comprises wheel-IMU-GNSS-odometry (WIGO), odometry-guided SfM reconstruction, and road surface reconstruction.
\subsubsection{WIGO}
\begin{figure}[t!]
    \includegraphics[width=0.48\textwidth]{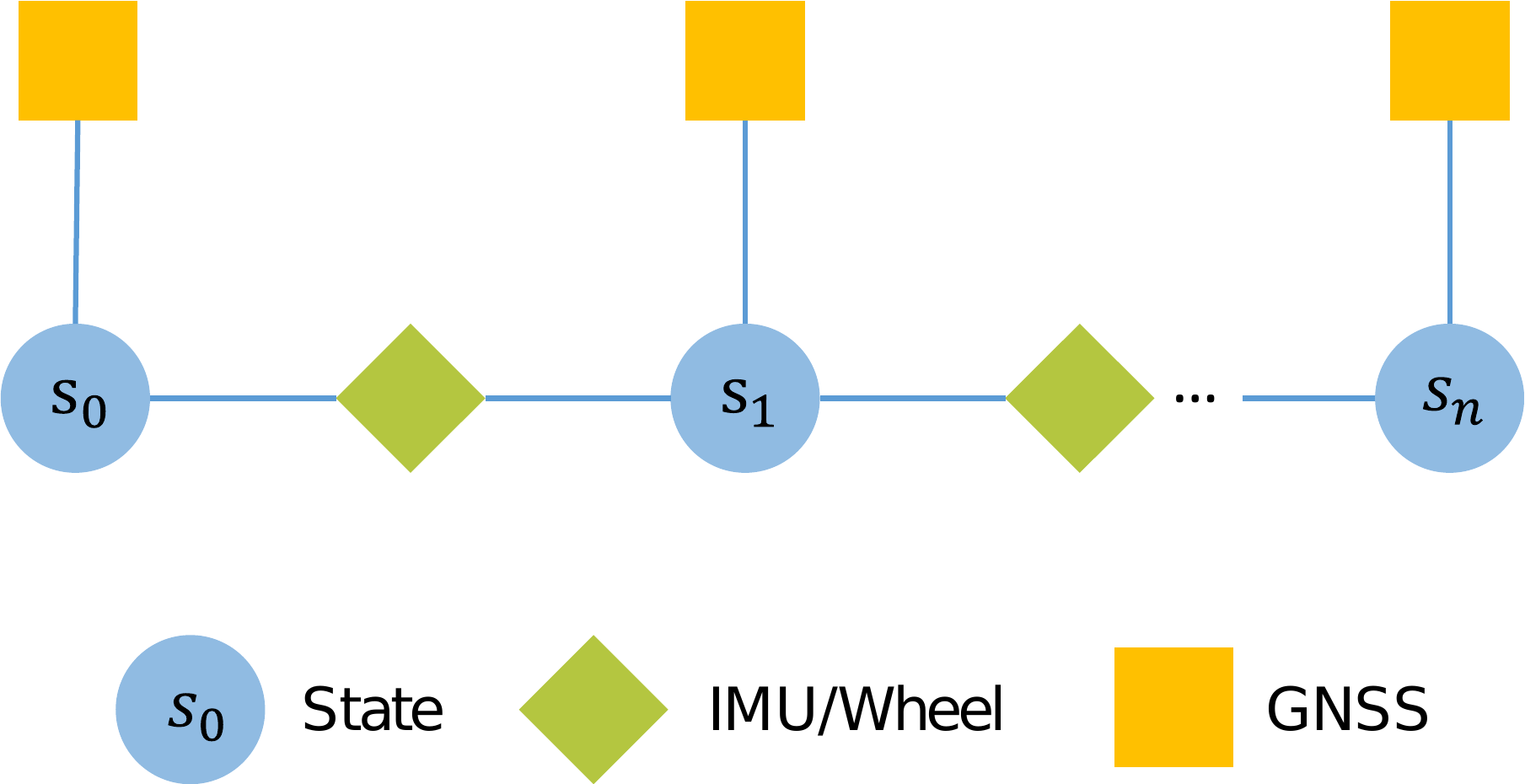}
  \caption{WIGO algorithm. GNSS, IMU, and wheel are fused in pose graph optimization to obtain accurate global poses.}
  \label{fig:wigo}
  \vspace{-1em}
\end{figure}

For locally accurate and globally drift-free pose estimation, multiple sensors with complementary properties are usually fused together~\cite{qin2019general}. Following VIWO~\cite{lee2020visual}, we propose a WIGO algorithm to combine the local sensor (wheel and IMU) with the global sensor (GNSS) in a pose graph optimization~\cite{gtsam, factor_graphs_for_robot_perception}. As depicted in Fig.~\ref{fig:wigo}, each pose state $\mathbf{s}$ serves as one node in the pose graph, which contains position and orientation in the world frame. The edge between two consecutive nodes is a local constraint from the IMU and wheel encoder. Another edge is a global constraint, which comes from the GNSS. Within graph optimization, local estimations are aligned with a global coordinate. Meanwhile, the accumulated drifts are eliminated. The WIGO algorithm gives coarse global 6-DoF (Degree of Freedom) poses with a real-world scale, and these results are used as inputs of the SfM pipeline illustrated below.

\subsubsection{Odometry Guided SfM Reconstruction}
\begin{figure}[t!]
    \includegraphics[width=0.48\textwidth]{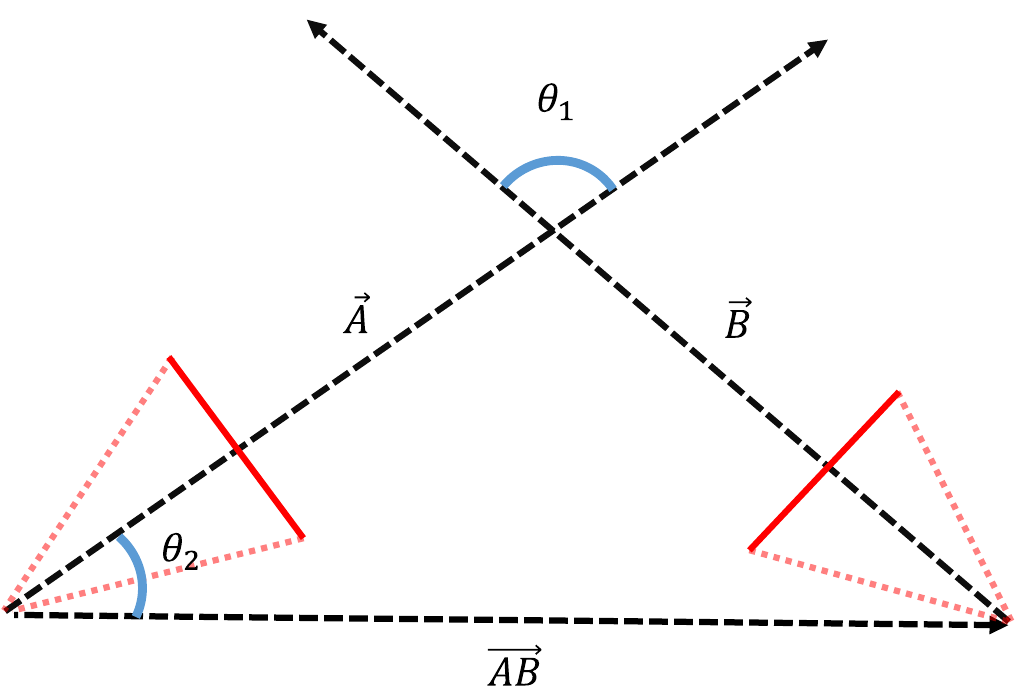}
  \caption{Illustration of homography-guided spatial pairs (HSP). The potential matching image pairs can be filtered by computing the visual overlap between different cameras, paying more attention to the road surface.}
  \label{fig:hsp}
  \vspace{-1em}
\end{figure}

We introduce an efficient SfM reconstruction method based on COLMAP~\cite{schoenberger2016sfm, schoenberger2016mvs} to achieve higher efficiency and accuracy. We optimize COLMAP from the following six aspects to meet the whole scene reconstruction toward intelligent driving:

\noindent \textbf{Homography-guided Spatial Pairs}: For a complete 3D scene reconstruction, more spatial points with accurate locations are needed. Thus, certain strategies are required to obtain dense matching points. In the case of unordered image data, exhaustive matching is frequently used, and the cost increases exponentially with the number of images. Sequential matching may alleviate this issue for sequential images captured by on-board cameras, but it introduces new problems with the lack of matching for multiple driving clips. We propose homography-guided spatial pairs (HSP) for cross-matching between multiple driving clips. HSP balance recall and efficiency of matching. Specifically, with the help of WIGO poses, we can obtain each camera's global pose as prior, the poses are defined as:
\begin{equation}
    \mathcal{S}=\{\mathbf{s}_{0},\mathbf{s}_{1},\ldots,\mathbf{s}_{n}\}.
\end{equation}

\begin{equation}
    \mathbf{s}_{i}=\{\mathbf{x}^{t}_{i},\mathbf{y}^{t}_{i},\mathbf{z}^{t}_{i},\mathbf{x}^{q}_{i},\mathbf{y}^{q}_{i}\,\mathbf{z}^{q}_{i}\,\mathbf{w}^{q}_{i}\}.
\end{equation}
each pose $\mathbf{s}$ consists of position $\mathbf{x}^{t}$, $\mathbf{y}^{t}$, $\mathbf{z}^{t}$ and 
orientation $\mathbf{x}^{q}$, $\mathbf{y}^{q}$, $\mathbf{z}^{q}$, $\mathbf{w}^{q}$. We use the k-nearest neighbours algorithm (KNN)~\cite{cover1967nearest} to find the nearest neighbour of each camera. For each pose $\mathbf{s}_{i}$ with its neighbour $\mathbf{s}_{j}$, it is added as a spatial pair if they are not far apart in the z-axis direction:
\begin{equation}
    \left|\mathbf{z}^{t}_{i}-\mathbf{z}^{t}_{j}\right|<\Delta\mathbf{z}^{t}.
\end{equation}
Then, we filter potential matching image pairs by computing the visual cone overlap between different cameras. As shown in Fig.~\ref{fig:hsp}, 
we can get the optical axis centre vectors $\overrightarrow{A}$ and $\overrightarrow{B}$ of the two cameras from the orientation of the cameras. By calculating the cosine similarity between the optical axes vectors $\overrightarrow{AB}$ and the cosine similarity between the vector and the vector $\overrightarrow{A}$, we can get the visual cone overlap of these two cameras:
\begin{equation}
    \cos\theta_{1}={\frac{\overrightarrow{A} \cdot \overrightarrow{B}}{\|\overrightarrow{A}\|\cdot\|\overrightarrow{B}\|}},
    \
    \cos\theta_{2}=\frac{\overrightarrow{AB} \cdot \overrightarrow{A}}{\|\overrightarrow{AB}\|\cdot\|\overrightarrow{A}\|}.
\end{equation}

If $\theta_{1}$ \textgreater $90^{\circ}$  and $\theta_{2}$ \textgreater $90^{\circ}$, which means that there is no visual cone overlap of the two cameras, so the spatial matching pair is filtered out. We also filter matching pairs where the camera's optical axis direction is face-to-face and close together. Image pair matching can be misleading due to the proximity and unreasonable viewing angle. Furthermore, all the cameras have an approximate extrinsic to the ground plane for intelligent driving applications. By applying the homography transformation to the ground plane, the visual overlap between cameras can be further filtered to emphasize the importance of the road surface area.

\begin{figure*}[t!]
    \includegraphics[width=1\textwidth]{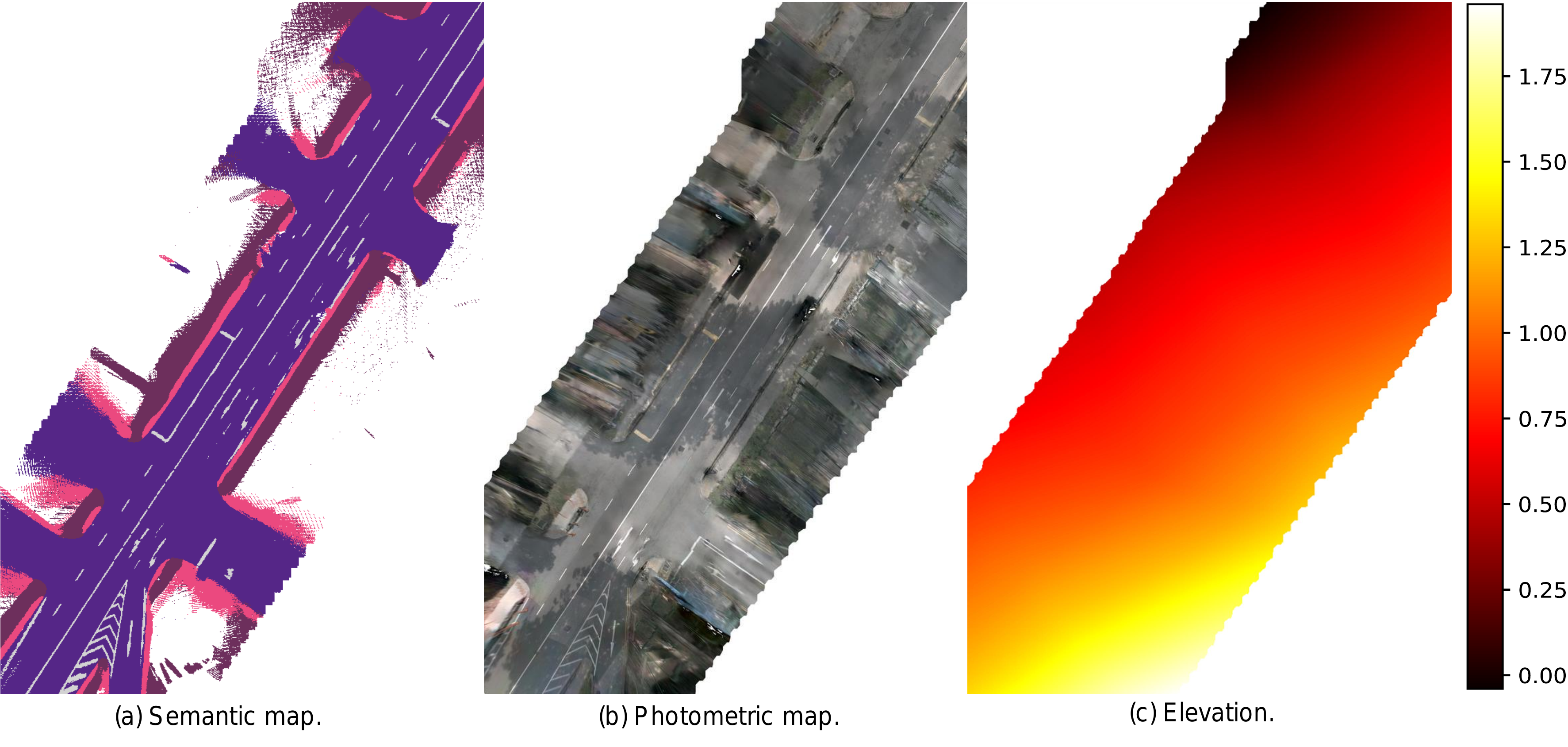}
    \caption{Using our proposed method, we reconstruct an HD map of a $300m \times 300m$ site with multiple scenes in nuScenes. (a) Semantic map in BEV, purple, pink, and white correspond to road surface, road teeth, and lane marking, respectively. (b) Photometric map in BEV. (c) Elevation visualization in hotmap, brighter indicates higher.}
    \label{fig:bev_result}
  % \vspace{-1em}
\end{figure*}

\noindent \textbf{Deep Learning Based Correspondence Search}: To further improve the robustness of our pipeline in the event of insufficient illumination and extreme weather conditions, we train a feature point extraction network, SuperPoint~\cite{detone2018superpoint}, on our driving dataset and pay extra attention to the weakly-textured or textureless road surface. In addition, the matching task encounters considerable challenges for images with significant variations in viewpoint and lighting conditions. We use SuperGlue~\cite{sarlin2020superglue} for local feature matching to improve the robustness and accuracy of matching in complex scenes.

\noindent \textbf{Odometry Guided Initialization}: Initialization is a crucial step in SfM, which directly affects the robustness, accuracy, and performance of reconstruction. Incremental SfM computes the scene graph for a robust reconstruction process to find the best initialization. However, this initialization may suffer substantial computation burdens when dealing with large-scale reconstruction (e.g., $300m \times 300m$ areas and thousands of images). In real-world driving scenarios, by fusing multiple sensors (e.g., GNSS, IMU and wheel) and applying localization algorithms, the ego vehicle poses information can be obtained, from which the position and orientation of each camera can be determined. Inspired by this, we propose an odometry-guided initialization (OGI) for SfM. At the beginning of the reconstruction process, the WIGO pose is transformed into the camera coordinates with extrinsic. Given the initial poses, the incremental SfM can be replaced with the spatial-guided SfM, and the real-world scale camera pose initialization dramatically speeds up the reconstruction process.

\noindent \textbf{Parallel Reconstruction}: Directly reconstructing all the images in a large-scale scene is not recommended. The main reason is that it not only exceeds the memory capacity of a single computer but also makes it difficult to take full advantage of parallel computing. A large-scale reconstruction area commonly consists of multiple driving clips. We, therefore, reconstruct the driving clips individually as a unit, with each clip usually having hundreds of images, and then clips with enough visual cone overlap are merged according to HSP, finally obtaining a complete large-scale reconstruction. In practice, we reconstruct each clip parallel to OGI, followed by triangulation and bundle adjustment (BA). We perform re-triangulation and global BA on the merged model to filter outliers and improve reconstruction results.

\noindent \textbf{Iterative BA}: After triangulation, the first BA is severely affected by outliers, and a second BA step can significantly improve the results. Thus, we propose an iterative BA strategy, using the pre-BA results for re-triangulation, followed by an optimization step on the post-BA results. The inaccurate points can be removed from the SfM sparse model step by step so that the accuracy and robustness of the optimization results can be significantly improved. We provide an overview of our iterative BA in Algorithm~\ref{alg:iter_BA}. In the majority of cases, it is after the third iteration that we see a significant leap in the results. This is when the optimization converges, marking a key point in the process.

\begin{algorithm}[t]
\caption{Iterative Bundle Adjustment}\label{alg:iter_BA}
\begin{algorithmic}
\STATE 
\STATE {\textbf{Input:}}
\STATE \hspace{0.5cm}All reconstructed points, $X$;
\STATE \hspace{0.5cm}All camera parameters, $P$;
\STATE \hspace{0.5cm}All images, $N$;
\STATE {\textbf{Output:}}
\STATE \hspace{0.5cm}Refined points and camera parameters;
\STATE \hspace{0.5cm}Initialize iteration number: $i \xleftarrow{} 1$;
\STATE \hspace{0.5cm}\textbf{While} $i \leq max\ iteration$ \textbf{do}
\STATE \hspace{1.0cm}$\Tilde{X_{i}}, \Tilde{P_{i}}, \Tilde{N_{i}} \xleftarrow{} Retriangulation(X_{i}, P_{i}, N_{i})$;
\STATE \hspace{1.0cm}$O_{i} \xleftarrow{} ComputeNumObservations(\Tilde{X_{i}}, \Tilde{P_{i}}, \Tilde{N_{i}})$;
\STATE \hspace{1.0cm}$\hat{X_{i}}, \hat{P_{i}}, \hat{N_{i}} \xleftarrow{} postBARetriangulation(\Tilde{X_{i}}, \Tilde{P_{i}}, \Tilde{N_{i}}) $; 
\STATE \hspace{1.0cm}$C_{i} \xleftarrow{} FilterPoints(\hat{X_{i}}, \hat{P_{i}}, \hat{N_{i}})$;
\STATE \hspace{1.0cm}$R_{i} \xleftarrow{} C_{i} / O_{i}$;
\STATE \hspace{1.0cm}\textbf{if} $R_{i} \leq max\ refinement\ change$ \textbf{then}
\STATE \hspace{1.5cm}\textbf{break}
\STATE \hspace{1.0cm}$FilterImagesWithReprojError(\hat{X_{i}}, \hat{P_{i}}, \hat{N_{i}})$;
\STATE \hspace{1.0cm}$i \xleftarrow{} i+1$
\end{algorithmic}
\label{iter_BA}
\end{algorithm}

\noindent \textbf{Rigid Prior}: For intelligent driving applications, multiple cameras attached to the vehicle can be regarded as mounting on a rigid body. It means that the position and orientation between them are relatively fixed, so we do not need to optimize each camera's position and orientation as independent variables. In the ordinary BA process, we need to optimize all the camera parameters $P$ and point parameters $X$, which are described by the following equation:
\begin{equation}
    \label{equ:ordinBA}
    \min_{P_i,X_j}\sum_{i=1}^N\sum_{j\in\mathcal{O}(i)}\|u_{ij}-\pi\big(P_i,X_j\big)\|^2.
\end{equation}
where $u_{ij}$ is an observed image point, $\pi(P_i,X_j)$ is the projection of point $X_{j}$ on camera $i$ with camera parameter. Since the relative positions and orientations between the cameras are fixed, we propose to use a rigid BA instead of the ordinary BA. In this case, we optimize a global rigid transformation to adjust all camera poses. We can rewrite the Equation~\ref{equ:ordinBA} by:
\begin{equation} 
    \label{equ:rigidBA}
    \min_{T,X_{j}}\sum_{i=1}^{N}\sum_{j\in\mathcal{O}(i)}\|u_{ij}-\pi\big(TP_{ref}R_{i},X_{j}\big)\|^{2}.
\end{equation}
where $T$ is the global rigid transformation matrix, describing the position and rotation of the entire rigid body, $P_{ref}$ is the pose parameter of the reference camera, $R_{i}$ is the fixed relative transformation matrix of camera $i$ with respect to the reference camera.
Cameras with large relative position changes are considered incorrectly estimated and filtered after rigid BA. The application of rigid BA not only enables more efficient processing of multi-camera systems but also improves the accuracy of the reconstruction results.
% \end{itemize}

With the above optimizations, we achieve roughly five times efficiency boost and 20\% robustness (success rate) improvements for intelligent driving datasets (see Section~\ref{sec:exp}. The accurate 6-DoF poses and corresponding sparse 3D points generated by SfM models are used as input for RoMe to reconstruct road surface mesh (Section~\ref{s:approach:rome}).

\subsubsection{Road Surface Reconstruction}
\label{s:approach:rome}
We extend our previous work RoMe~\cite{mei2023rome} for road surface mesh reconstruction. The mesh reconstruction can be roughly divided into three tasks:

% \begin{itemize}
\noindent \textbf{Surface Points Initialization}: To reduce the impact of moving objects and parked vehicles on road surface reconstruction, we apply an off-the-shelf 2D segmentation network~\cite{cheng2022masked} to remove occluding objects and obtain lane segmentation masks. Combined with the sparse SfM models, semantic sparse point clouds can be recovered. The sparse road surface point clouds are then extracted. To further improve the robustness when the SfM points are too sparse, we initialize a narrow road surface based on ego-pose and sample the surrounding SfM points. Ultimately, a larger range of road surface points is obtained. This approach improves the overall quality of the road surface initialization.

\noindent \textbf{Elevation Estimation}: Using the sparse road surface obtained in the previous step, we trained an elevation multi-layer perceptron (MLP)~\cite{gardner1998artificial} for dense road elevation prediction. Unlike RGB and semantic information, the road elevation change does not change drastically, and we use position encoding to control the smoothness of the road surface. 

\noindent \textbf{Mesh Optimization}: A mesh is initialized based on the predicted dense road elevation. The original image and its corresponding 2D segmentation results are used for training to assign semantic labels and photometric features to each triangular facet in the mesh.

In practice, the elevation MLP is optimized and refined during the mesh optimization stage to improve the consistency between geometry and photometric features. Eventually, the highly accurate 3D road surface mesh can be obtained. Notably, for ease of annotation, the 3D road surface mesh can be represented as 2D BEV images and an elevation map, as depicted in Fig.~\ref{fig:bev_result}. Such representation is fed into the next stage: Map Annotation (Section~\ref{s:approach:annotation}).

\begin{figure}[t!]
    \includegraphics[width=0.48\textwidth]{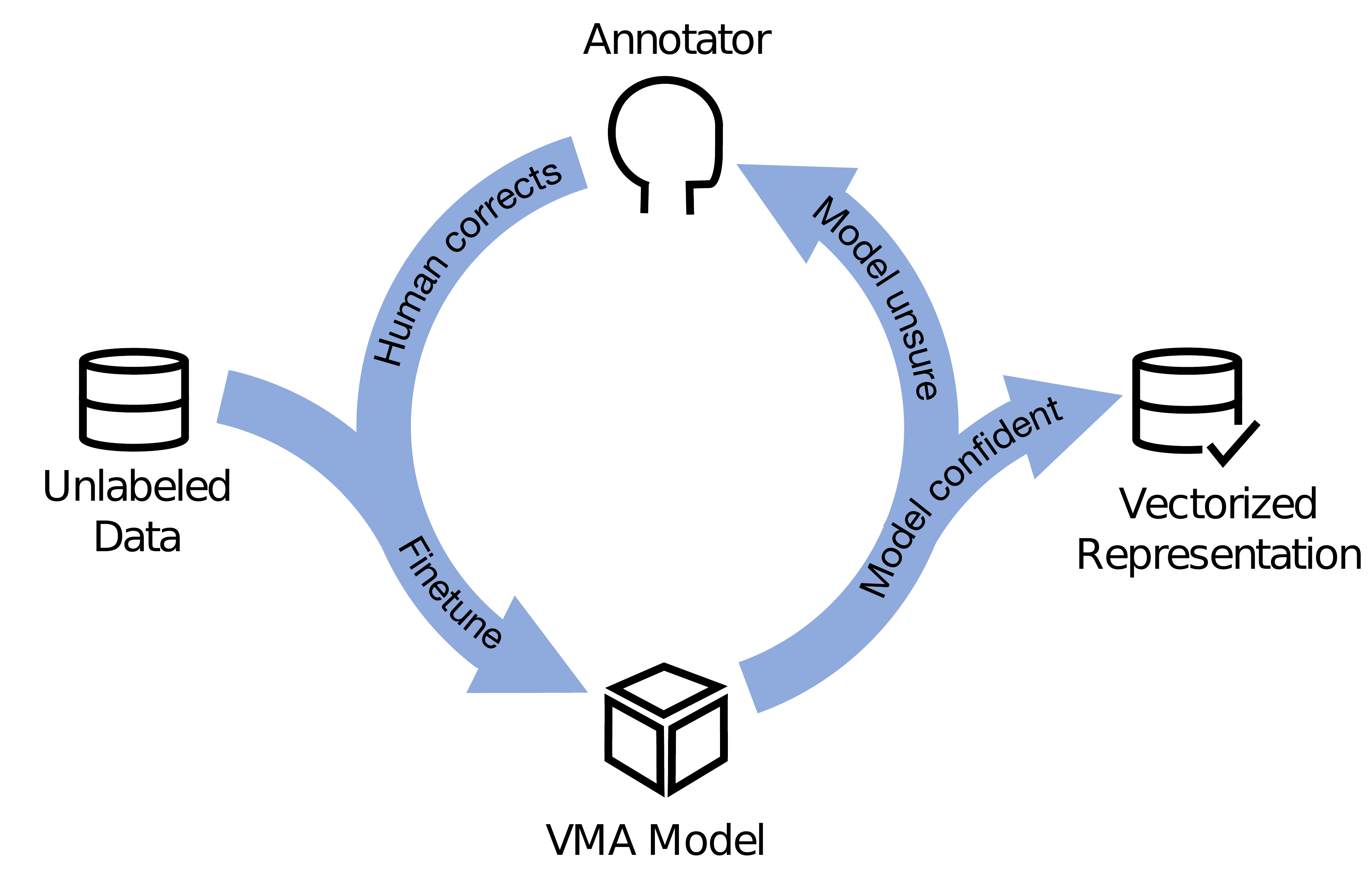}
  \caption{``Human in the loop" mechanism in HD map annotation. The VMA results are used as priors to accelerate the manual annotation process. While the labelled data is accumulated, the manual works are converted from labelling to verification and minor modification.}
  \label{fig:hitp}
  \vspace{-1em}
\end{figure}

\subsection{Map Annotation}
\label{s:approach:annotation}
We propose a semi-automated approach based on human-in-the-loop for accelerating time-consuming HD map annotation. As shown in Fig.~\ref{fig:hitp}, trained annotators will manually label all the data and generate HD maps in the beginning. Once enough labelled data is collected, a neural network is trained with supervised learning to annotate HD maps automatically. During this process, we keep the high-confidence outputs of the model, relabel the low-confidence ones, and the human-annotated results are fed back to retrain the model in the next iteration. In particular, we extend the vectorized map annotation (VMA) system as a learnable annotation model.
% to overwrite
VMA is an automatic offline map annotation framework based on MapTR~\cite{liao2022maptr}. The input is  2D BEV images, while the output is the vectorized representation of road surface elements (e.g., lane dividers, road boundaries, pedestrian crossing). We propose to use the concatenated 2D BEV semantic photometric images as inputs since they can improve the VMA reasoning ability, especially when classifying the type of lane dividers. Note that this process is still in 2D BEV space. After obtaining the vectorized 2D representations, an elevation map is combined to lift the 2D vectors into actual 3D vectors.

In the actual annotation process, the VMA model is constantly optimized with the accumulation of annotation data. The manual works are converted from the initial annotation to verification and fine-grained modification. The human-in-the-loop mode greatly accelerates the annotation efficiency.

\begin{figure*}[t!]
    \includegraphics[width=1\textwidth]{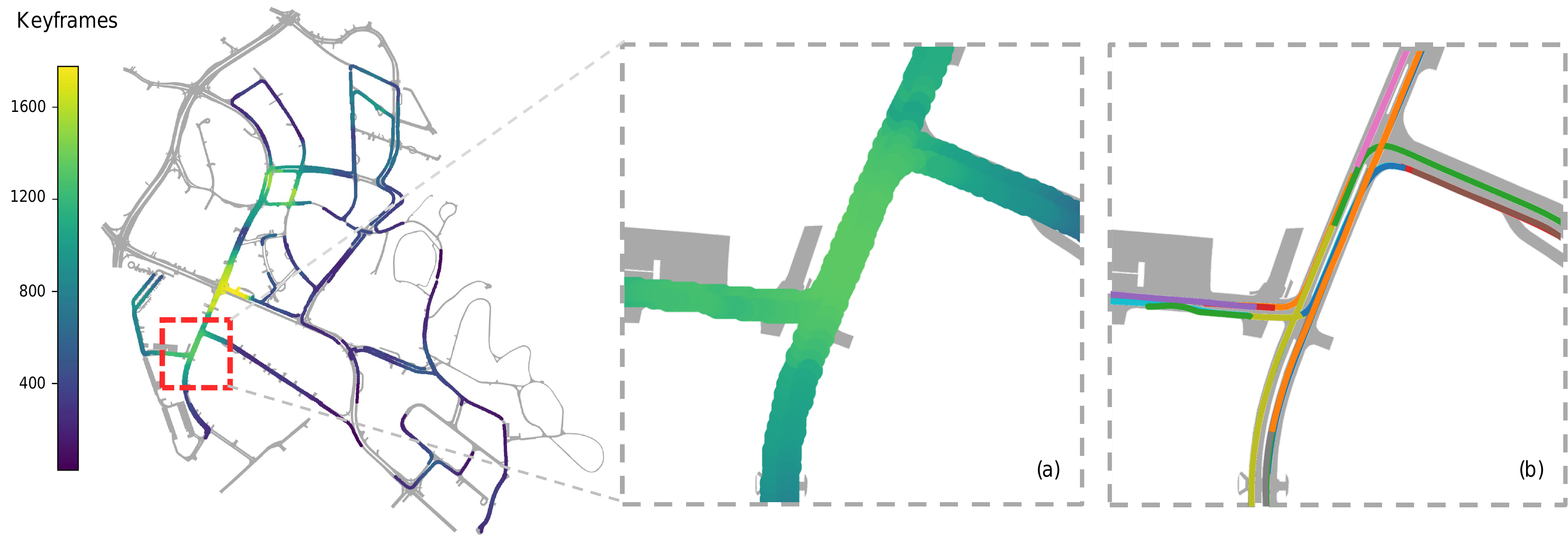}%
    \caption{Spatial data coverage for one nuScenes location (Singapore one north). The colour bar indicates the number of keyframes with ego vehicle poses within a 100m radius across all scenes. (a) Keyframe distribution of a crossroads. (b) Trajectories of different scenes (with different colours) at the crossroads. Best viewed in colour.}
    \label{fig:show_site}
\end{figure*}

\section{Experiment}
\label{sec:exp}
We first introduce the dataset used to validate our proposed framework and, in this way, generate optimized annotations. Then, we evaluate the proposed CAMAv2 through a set of quantitative and qualitative comparisons, including the state-of-the-art online HD map reconstruction methods. Finally, we do not shy away from discussing the long-tail cases and dirty details behind the CAMAv2 in the last part.

\subsection{Data Samples}
The nuScenes dataset~\cite{caesar2020nuscenes} is one of the most widely used The nuScenes dataset~\cite{caesar2020nuscenes} is one of the most widely used datasets for online HD map construction~\cite{li2022hdmapnet, liu2023vectormapnet, liao2022maptr, liao2023maptrv2,chen2022persformer}. It contains 1000 scenes of 20 seconds each, all taken from four different locations. There are 28130, 6019, and 6008 key samples for training, validation, and testing. All of these key samples are synchronized using LiDAR keyframes. Due to the inherent flaws in the road surface annotation pipeline, the nuScenes HD map lacks elevation information. Thus, we validate our proposed framework on the nuScenes dataset to generate and publicly release optimized annotations.

\subsection{Site Reconstruction}
Fig.~\ref{fig:show_site} (far left) presents the spatial distribution of the keyframe of all scenes on a city-scale map. It can be observed that the majority of the data comes from intersections, indicating that most of the scenes in the nuScenes dataset have geographic overlap. As the map is zoomed in, it becomes evident that multiple scenes traverse the same location and make multiple observations (see Fig.~\ref{fig:show_site} (a, b)). For improvement, we propose a multi-scene aggregation reconstruction method. This reconstruction method aggregates scenes with intersecting portions into one large scene called a \textbf{site}. Site reconstruction utilizes more observations to reconstruct the same location, which solves the shortcoming of dropping the head and tail frames in the previous single-scene reconstruction and the occlusion and blind zone problems~\cite{zhang2024cama}. In the final annotation stage, we work on the site-reconstructed map only once to get each scene's annotation. It can ensure the completeness of the local map reconstruction, meet the needs of long-distance ground-truth perception, accelerate annotation efficiency, and reduce costs. We first use the ego-pose provided by nuScenes for site aggregation. The ego pose of the aggregated site is then converted to each camera's pose for SfM initialization. For self-acquisition data, the WIGO localization algorithm can generate precise poses to guide SfM reconstruction.

In practice, we ignore the nighttime scenes during the aggregation process. This is because most of the nighttime scenes in nuScenes datasets have poor lighting conditions, the ground is not visible, and aggregating these scenes with other daytime scenes affects the reconstruction results. To fairly compare our proposed annotations with the original ones, we subsample the nuScenes dataset according to our annotation frames, namely nuScenes-sub and nuScenes-CAMAv2.

\begin{figure}[t!]
    \includegraphics[width=0.48\textwidth]{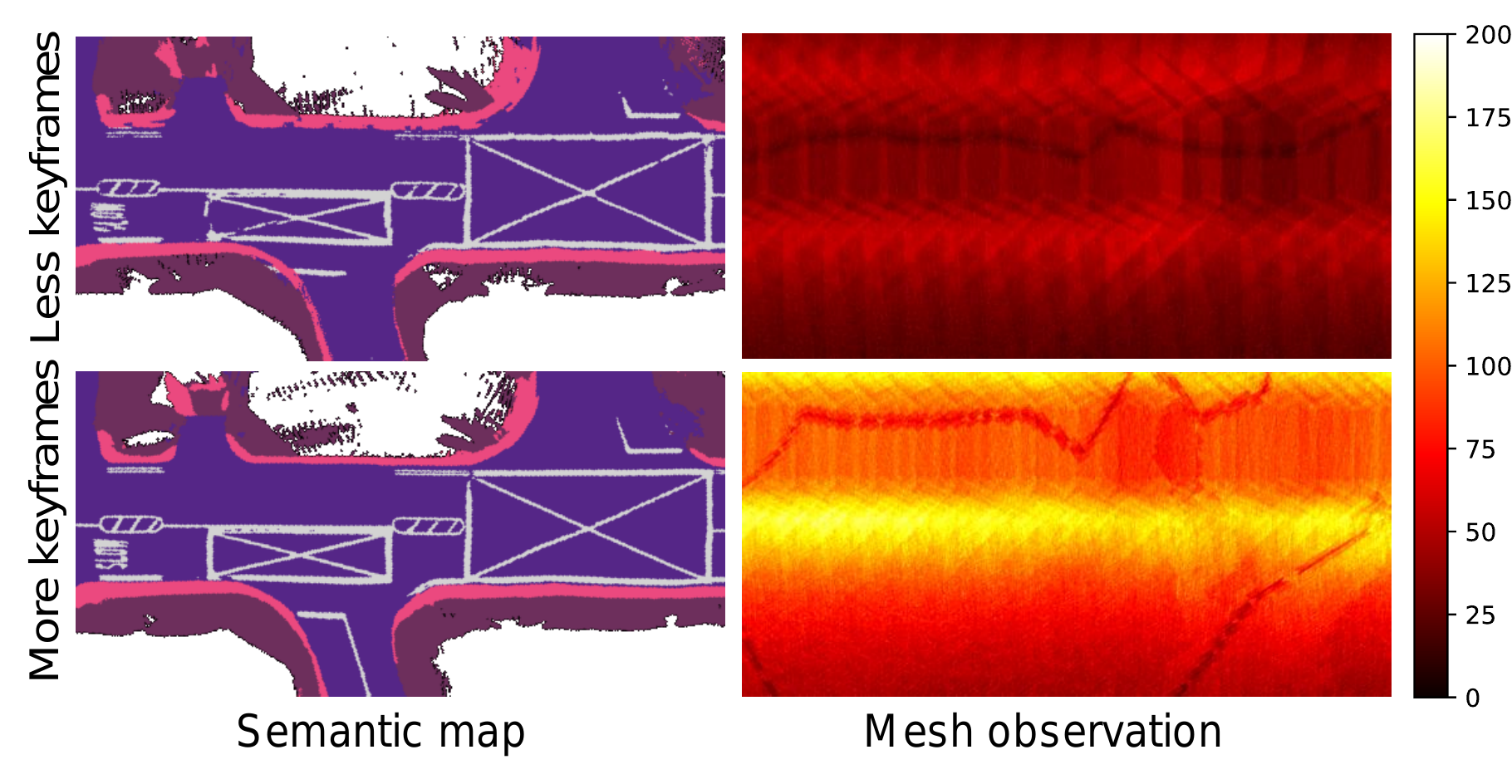}
  \caption{Mesh observation distribution of one reconstructed road surface. The left column shows the reconstructed semantic map and the right column shows the number of times all camera rays hit each vertex.}
  \label{fig:mesh}
  \vspace{-1em}
\end{figure}

\subsection{Annotation Properties}
Following the convention of existing online HD map reconstruction methods, three types of map elements are annotated: pedestrian crossing, lane divider, and road boundary. We have undertaken a comprehensive comparison between the original nuScenes-sub annotations and our proposed nuScenes-CAMAv2 annotations. This comparison is based on the number of different category annotations in the training and validation sets, as detailed in Table~\ref{tab:anno_prope}.

\begin{table}[h]
\normalsize
\centering
    \caption{Statistics of the static map element number.}
    \begin{tabular}{c c c c}
    \hline
                & Lane divider  & Ped. crossing  & Road boundary  \\ \hline
    nuScenes    & 175665        & 44073    & 119191    \\ \hline
    CAMAv2        & 141745       & 31976    & 137941    \\ \hline
    \end{tabular}
    \label{tab:anno_prope}
\end{table}

\subsection{Quantitative Validation}
We first report the time cost of each component in CAMAv2. After that, we quantitatively evaluate the accuracy and consistency of CAMAv2 annotation with the original ones, and finally quantify how many observations are required in road surface reconstruction.

% \begin{itemize}
\noindent \textbf{Time-Costing:} A comparison of the average time cost before and after the improvement for each component is detailed in Table~\ref{tab:runtime}, which counts the time cost for reconstructing a $300m \times 300m$ site (around eight clips and three thousand frames per site).
We find feature matching and SfM reconstruction is the most time-consuming part. The HSP and OGI SfM greatly accelerate the reconstruction process. Annotation of the static map elements using the 2.5D method is more efficient than 3D annotation. With continuous fine-tuning of the VMA model, the final annotation efficiency can be improved more than two times.

\noindent \textbf{Consistency and Accuracy:} As introduced in Section~\ref{sec:intro}, the nuScenes dataset provides HD maps for all scenes, while the annotation does not contain elevation. Reprojecting an HD map into the image space using the ego poses and calibration, as presented in Fig. 1, we can clearly observe the misalignments between the lane vectors and image space. For quantitative analysis, we propose a metric denoted semantic reprojection error (SRE) to analyze better and compare the reprojection accuracy. The detailed steps are illustrated in Algorithm~\ref{alg:sre}: 
\begin{itemize}
    \item \textbf{Step 1:} Reproject the 3D annotation vector elements into each 2D image.
    \item \textbf{Step 2:} Extract all the instances in image space using an off-the-shelf 2D lane instance segmentation~\cite{jain2023oneformer, cheng2022masked, kirillov2023segment} and fit polylines for each instance.
    \item \textbf{Step 3:} Match the elements projected in step 1 and the elements extracted in step 2 using the Hungarian algorithm~\cite{kuhn1955hungarian}.
    \item \textbf{Step 4:} Calculate the mean pixel distance for each matched element.
\end{itemize}

\begin{table}[t!]
\setlength{\tabcolsep}{16pt}
\normalsize
\centering
    \caption{Time cost (on average) of each CAMAv2 component.}
    \begin{tabular}{c c c c}
    \hline
    Corresp. Search    & SfM R.C.       & Surface R.C.  \\ \hline
    90 mins            & 140 mins      & 120 mins         \\ \hline
    \end{tabular}
    \label{tab:runtime}
\end{table}

\begin{table}[t!]
\normalsize
\centering
   \caption{Quantitative comparison between the original nuScenes HD map and our proposed nuScenes-CAMAv2 HD map. We evaluate the Semantic Reprojection Error (SRE), Precision, and Recall.}
   \begin{tabular}{c c c c c}
   \hline
            &  SRE $\downarrow$ & Precision $\uparrow$ & Recall $\uparrow$  & $F_1 \uparrow$ \\ \hline
   nuScenes &  8.03           & 0.59               & 0.51             & 0.54               \\ \hline
   CAMAv2 (Ours)   &  \textbf{4.96}  & \textbf{0.89}      & \textbf{0.57}    & \textbf{0.69}      \\ \hline
   \end{tabular}
   \label{table:comparison}
\end{table}

\begin{algorithm}[t]
\caption{Semantic Reprojection Error}\label{alg:sre}
\begin{algorithmic}
\STATE 
\STATE {\textbf{Input:}}
\STATE \hspace{0.5cm}All camera poses, $T$;
\STATE \hspace{0.5cm}All camera intrinsics, $K$;
\STATE \hspace{0.5cm}2D lane instance segmentation, $L$;
\STATE \hspace{0.5cm}HD maps 3D vectors, $M$;
\STATE {\textbf{Output:}}
\STATE \hspace{0.5cm}Semantic Reprojection Error, $SRE$;
\STATE \hspace{0.5cm}\textbf{for} frame $i$ in range(\# of images) \textbf{do}
\STATE \hspace{1.0cm}transform HD map from world to camera frame;
\STATE \hspace{1.0cm}Map in each frame $M_{i} \xleftarrow{} transform(T_{i},M)$;
\STATE \hspace{1.0cm}Crop $M_{i} \xleftarrow{} crop(M_{i},x_{max},x_{min},y_{max},y_{min})$;
\STATE \hspace{1.0cm}Project HD map $M_{i} \xleftarrow{} project(K,M_{i})$;
\STATE \hspace{1.0cm}Match $M_{i}$ and $L_{i}$ with Hungray algorithm;
\STATE \hspace{1.0cm}Matched pairs $P \xleftarrow{} \{M_{ij},L_{ij}\}$;
\STATE \hspace{1.0cm}\textbf{for} instance $j$ range(\# of $P$) \textbf{do}
\STATE \hspace{1.5cm}$L_{ij} \xleftarrow{} skeletonize(L_{ij})$;
\STATE \hspace{1.5cm}\textbf{for} pixels $S_k$ in $L_ij$ \textbf{do}
\STATE \hspace{2.0cm}$d_{k} \xleftarrow{} points\ to\ curve\ distance(S_{k},M_{ij})$;
\STATE \hspace{1.5cm}\textbf{end for}
\STATE \hspace{1.5cm}$Error{M_{ij}, L_{ij}} \xleftarrow{} mean(d_k)$;
\STATE \hspace{1.0cm}\textbf{end for}
\STATE \hspace{1.0cm}$Error_{i}\xleftarrow{} mean(Error\{M_{ij},L_{ij}\})$;
\STATE \hspace{0.5cm}\textbf{end for}
\STATE \hspace{0.5cm}$SRE \xleftarrow{} mean(Error_{i})$
\end{algorithmic}
\label{sre}
\end{algorithm}

\noindent \textbf{Mesh Observation:} During training, each vertex is optimized by multiple images from different views. After optimizing all vertex (from thousands to millions according to mesh resolution), the final mesh (with elevation, colours, and semantics) is obtained to represent the whole road surface. Our research focuses on determining the optimal number of observations required to reconstruct the road surface effectively while maintaining a balance between performance and efficiency. We achieve this by controlling the density of keyframes involved in the reconstruction. To evaluate the amount of data needed for initialization, we count the number of times each vertex is observed from different viewpoints and visualize this data in a heatmap. As depicted in Fig.~\ref{fig:mesh}, the clarity of the reconstructed road surface marks improves with the increase in the number of observations, demonstrating the practicality of our approach.

\begin{table*}[t!]
\normalsize
\centering
   \caption{Comparison of different MapTRv2 model evaluation results on the nuScenes validation dataset. The training and validation use the same kind of annotation for each experiment.
   We also evaluate the reprojection accuracy (SRE) and consistency (precision, recall and $F_1$ score) for the prediction results of different models.}
   \begin{tabular}{c|c c |c c c c |c c c c c c}
   \hline 
    Exp. \#   &  annotation  & w/ elevation    & $AP_{ped} $ & $AP_{divider} $ & $AP_{boundary} $ & $mAP$    & SRE $\downarrow$ & Precision $\uparrow$ & Recall $\uparrow$  & $F_1 \uparrow$  \\ \hline
    \# 1      &  nuScenes    &                 & 60.8        & 57.7            & 61.2             & 59.9      &  8.43            & 0.51                 &  0.37              &  0.42 \\ \hline
    \# 2      &  CAMAv2        &                 & 41.6        & 41.8            & 38.9             & 40.7      &  6.42            & 0.75                 &  0.54              &  0.63  \\ \hline
    \# 3      &  CAMAv2        &  \checkmark     & 36.8        & 40.9            & 36.2             & 37.9      &  5.62            & 0.91                 &  0.51              &  0.65  \\ \hline
   \end{tabular}
   \label{table:model_training}
\end{table*}

\subsection{Qualitative Validation}
% 重投影可视化
%  openlane v2
We observe that the HD map provided by nuScenes does not always reflect the actual environments captured by cameras. For example, as shown in Fig.~\ref{fig:reprojection} (a, b), the HD map annotations indicate the lane divider between the ego vehicle and bicycle lane, but in the actual camera image, there is no lane marking on the ground in the relevant area. This is because lane dividers in HD maps are constructed based on rules rather than visually apparent markings, as the main goal is to map the world rather than facilitate intelligent driving applications. Such inconsistency between the image and HD map annotations increases the difficulty of training.

\begin{figure}[t!]
    \includegraphics[width=0.48\textwidth]{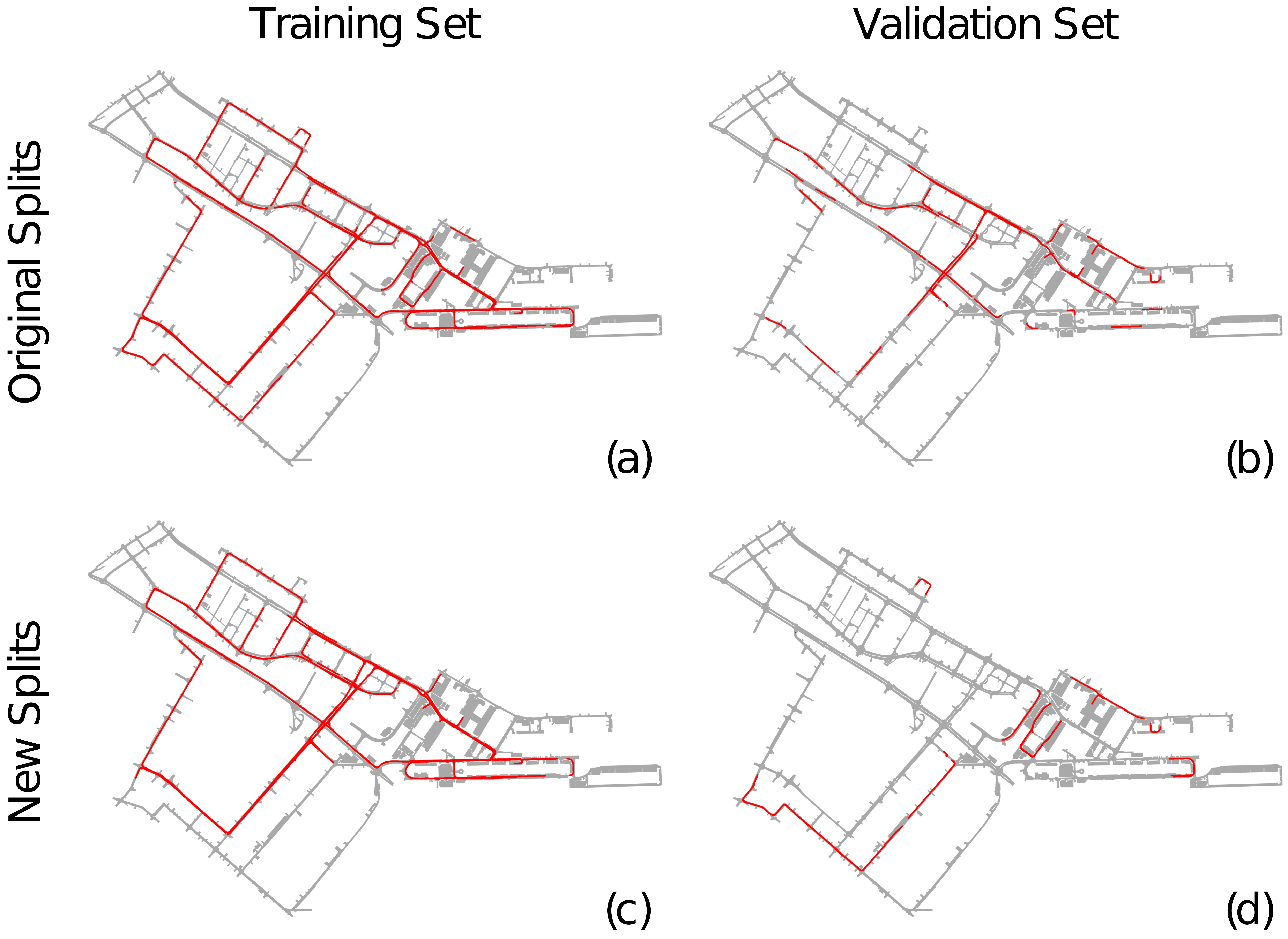}
  \caption{Different data splits for one nuScenes location (Boston seaport). The red lines indicate the spatial data coverage of training and validation sets. The original split has overlapping parts in the training set (a) and validation set (b), especially at central intersections. Our new split has no overlapping between the training set (c) and the validation set (d). The amount of training and validation data is the same after repartitioning.}
  \label{fig:splits}
  \vspace{-1em}
\end{figure}

An intuitive understanding: When the image and the HD map are inconsistent, the model is more likely to ``memorize" the HD map according to the surroundings instead of ``reasoning" the local map based on the observations. To verify this conjecture, we visualized the distribution of each scene in the nuScenes training and validation sets. Fig.~\ref{fig:splits} (a, b) shows that the original nuScenes dataset has overlapping parts in the training and validation sets. To verify whether duplicated observation validation data in the training set affects the model results, we repartitioned the training set and validation set based on the location distribution of each scene in the nuScenes dataset, and there is no overlap between the training set and validation set after our repartitioning (Fig.~\ref{fig:splits} (c, d)). We performed training and validation based on the newly segmented set. Table~\ref{table:model_new_split} shows that the model's accuracy on the new training data decreases substantially on the new validation data compared to the model obtained from the original data splits. It proves that the model is more inclined to memorize the HD map based on the surroundings rather than based on observations. It also thus side-steps the importance of highly accurate labelled data.
Consequently, this data inconsistency weakens the potential generalizability of the perception model. In contrast, CAMAv2 can generate accurate (Fig.~\ref{fig:reprojection} (d)) and consistent (Fig.~\ref{fig:reprojection} (f)) lane marking, including road boundary, lane divider, and pedestrian crossing.

\begin{table}[t!]
\normalsize
\centering
   \caption{Comparison of different MapTRv2 model evaluations on the nuScenes validation dataset. The training and validation use different data splits.}
   \begin{tabular}{c |c c c c}
   \hline 
   Data Split     & $AP_{ped}$  &$AP_{divider}$   &$AP_{boundary}$   &$mAP $  \\ \hline
    Original    & 59.8        & 60.2           & 61.5            & 60.5     \\ \hline
     New         & 26.4       & 14.4          & 37.9             & 26.2    \\ \hline
   \end{tabular}
   \label{table:model_new_split}
\end{table}

\begin{figure}[t!]
    \subfloat[MapTRv2 (trained with nuScenes) prediction.]
    {
        \includegraphics[width=0.48\textwidth]{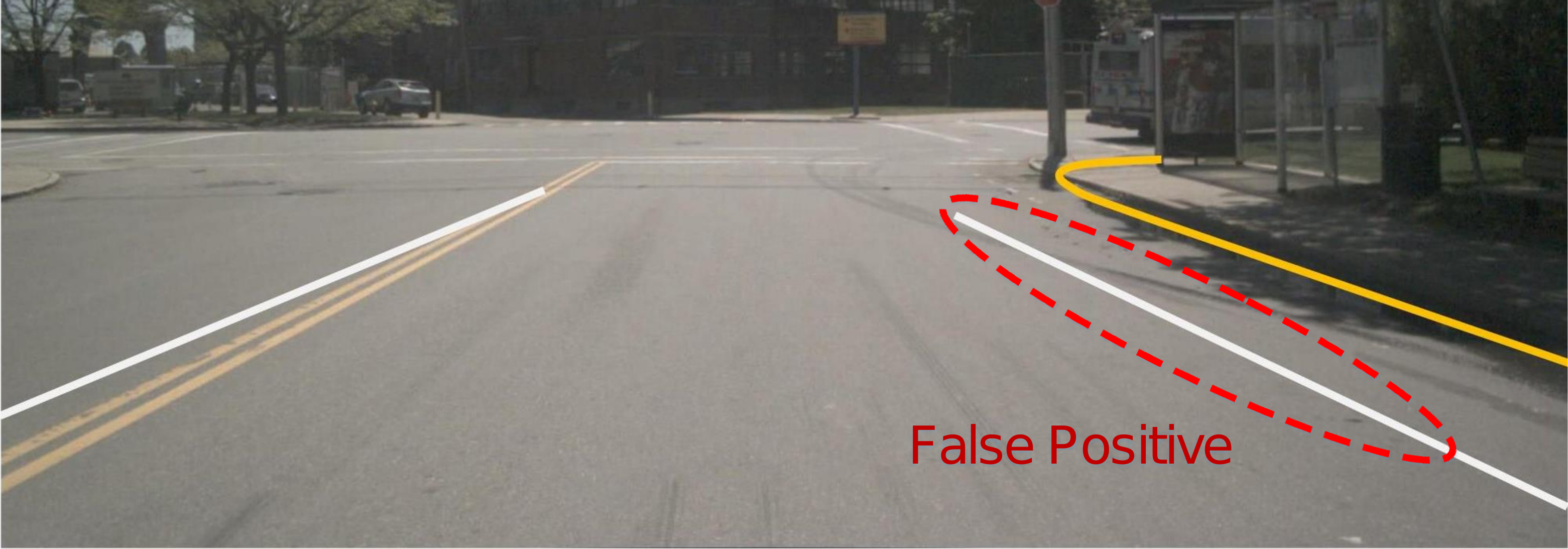}
    }
    \hfil
    \subfloat[MapTRv2 (trained with CAMAv2) prediction.]
    {
        \includegraphics[width=0.48\textwidth]{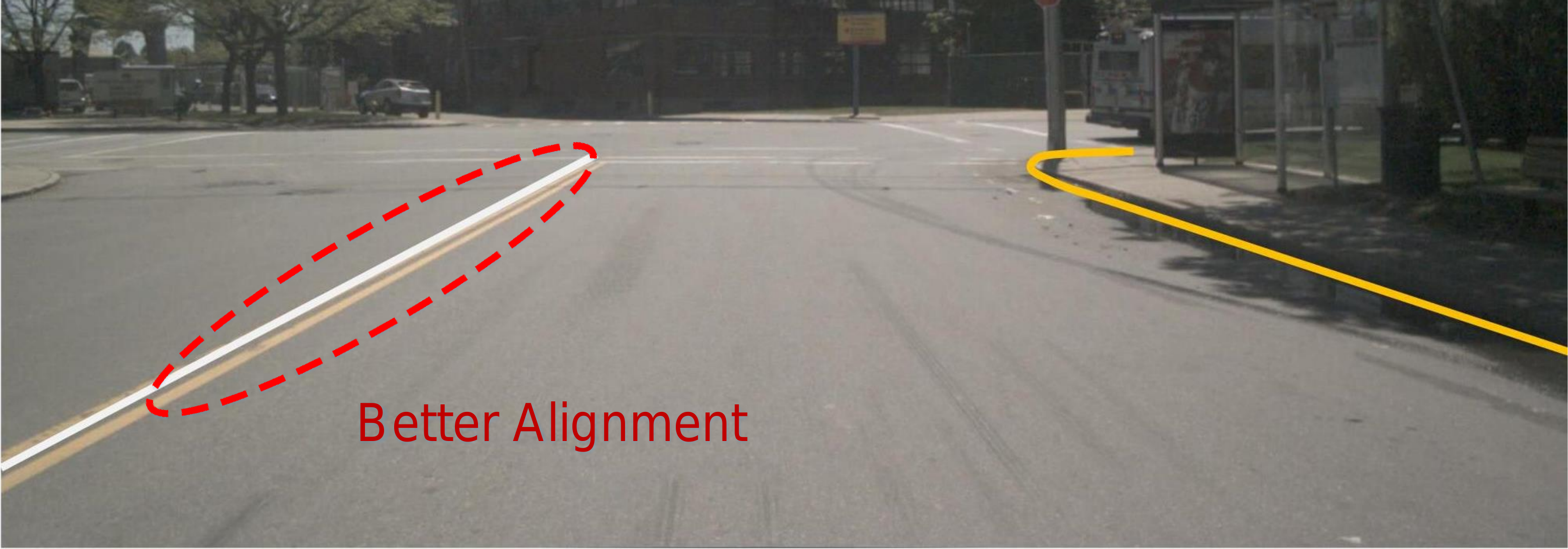}
    }
  \caption{Comparison of the model prediction reprojection. (a) The result reprojection of the MapTRv2 model trained with the nuScenes dataset. The red circle shows a false positive prediction, as there is no lane marking on the ground. (b) The result reprojection of the MapTRv2 model trained with the CAMAv2 annotation. The red circle shows better reprojection accuracy compared to (a).}
  \label{fig:model_result}
    \vspace{-1em}
\end{figure}

\subsection{Application}
High accurate and spatial-temporal consistent HD map annotations are vital for training BEV perception algorithms. To verify the effectiveness of our annotations, we choose MapTRv2 as the baseline model. We removed the auxiliary dense prediction loss in MapTRv2 to reduce the influence of irrelevant supervision signals. We conduct three experiments: 1) MapTRv2 trained on nuScenes-sub dataset; this is set as the baseline. 2) MapTRv2 trained on CAMAv2 but without elevation information. 3) MapTRv2 trained on the CAMAv2 annotation with elevation information. All the experiments are trained 24 epochs with ResNet-50 backbone~\cite{he2016deep}. Table~\ref{table:model_training} details the prediction accuracy and the reprojection metrics (including SRE and F1 score). We observe that models trained with CAMAv2 annotation predict higher reprojection accuracy and consistency. The SRE improves from 8.43 to 5.62 by 33 \%, and the $F_1$ score improves from 0.42 to 0.65.

We also visually compare the predictions in Fig.~\ref{fig:model_result}. Due to the inconsistent annotation in the nuScenes dataset, the trained MapTRv2 makes false positive predictions (red circle in Fig.~\ref{fig:model_result} (a)). With our proposed CAMAv2 annotation, the model predictions align better with the image in reprojection (red circle in Fig.~\ref{fig:model_result} (b)).

\subsection{Other Datasets}
\begin{figure*}[t!]
    \subfloat[Reconstruction result.]
    {
        \includegraphics[width=1\textwidth]{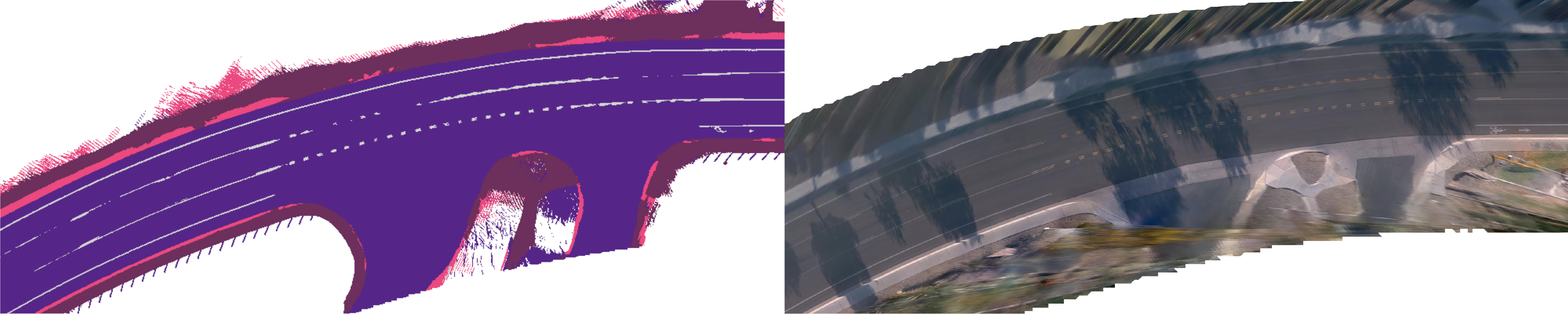}
    }
    \hfil
    \subfloat[Reprojection result.]
    {
        \includegraphics[width=1\textwidth]{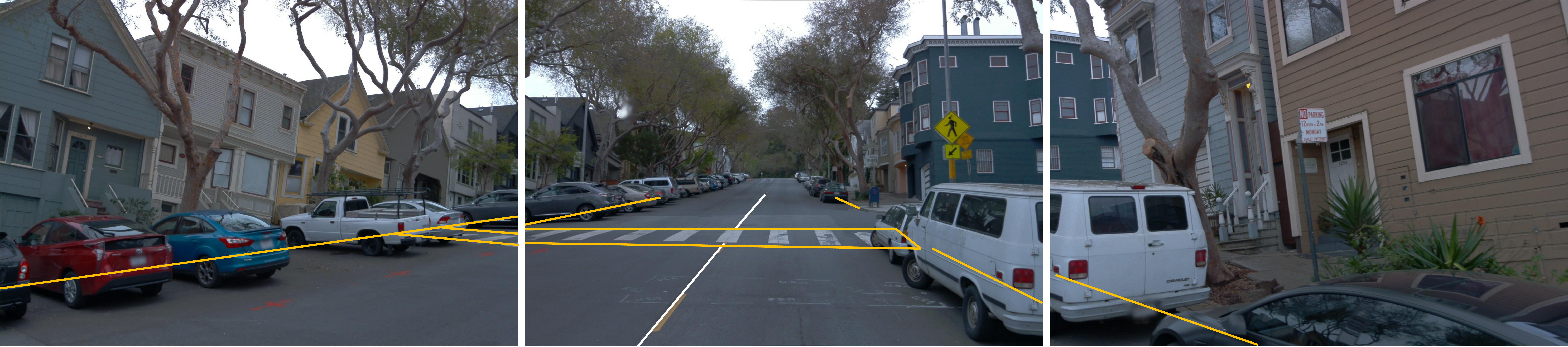}
    }
  \caption{Visualization and reprojection results in WOD. (a) Reconstructed road surface in BEV with semantic map on the left and photometric map on the right. (b) Reprojection results of WOD-CAMAv2 annotations in an uphill scenario.}
  \label{fig:waymo}
  \vspace{-1em}
\end{figure*}

To verify the robustness of our proposed CAMAv2, we validate it on other public datasets. Fig.~\ref{fig:waymo} (a) shows the effect of the reconstruction on the Waymo Open Dataset (WOD)~\cite{sun2020scalability}. We can accurately reconstruct ground elevation for the scenarios of uphill/downhill in WOD, and the reprojection results align well with the images (Fig.~\ref{fig:waymo} (b)). We also annotated several reconstructed WOD scenes and calculated the SRE. As shown in Table~\ref{table:waymo_sre}, the WOD-CAMAv2 annotations achieve low reprojection error.

\begin{table}[h]
\normalsize
\centering
   \caption{Quantitative analy on our proposed WOD-CAMAv2 HD map. We evaluate the Semantic Reprojection Error (SRE), Precision, and Recall.}
   \begin{tabular}{c c c c c}
   \hline
            &  SRE $\downarrow$ & Precision $\uparrow$ & Recall $\uparrow$  & $F_1 \uparrow$ \\ \hline
   CAMAv2 (Ours)   &  \textbf{2.77}  & \textbf{0.91}      & \textbf{0.73}    & \textbf{0.81}      \\ \hline
   \end{tabular}
   \label{table:waymo_sre}
\end{table}

\subsection{Long-tail Cases}
\begin{figure*}[t!]
    \subfloat[Rainy scenario.]
    {
        \includegraphics[width=1\textwidth]{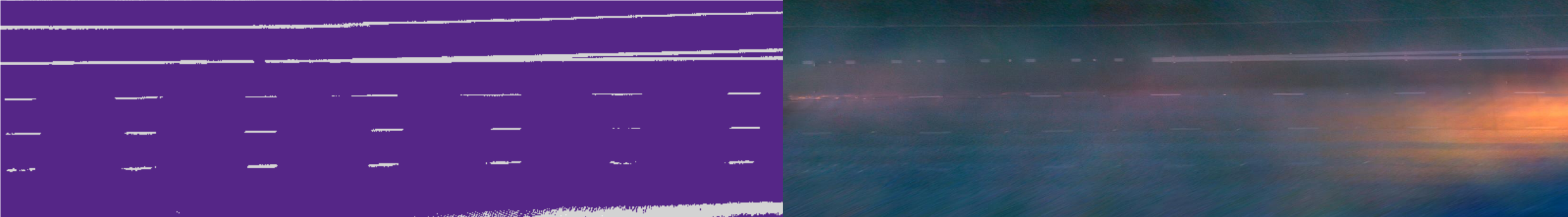}
    }
    \hfil
    \subfloat[Night scenario.]
    {
        \includegraphics[width=1\textwidth]{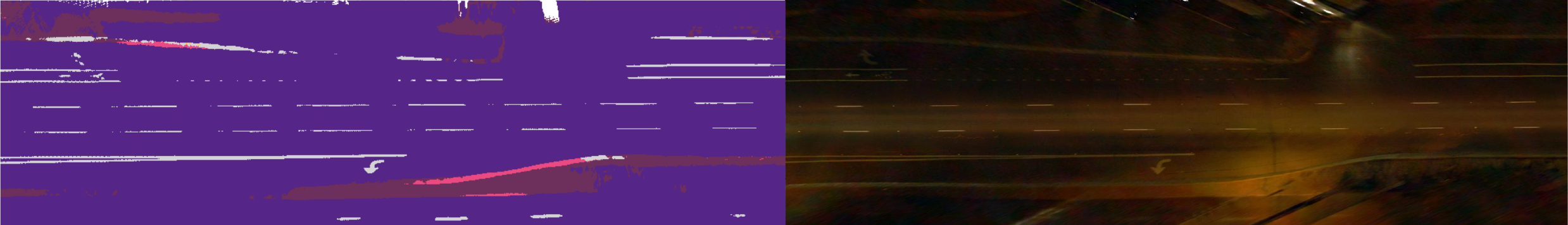}
    }
  \caption{Long-tail cases of our reconstructed results in WOD. (a) Reconstructed results in rainy weather, poor photometric map results due to lens soiling. (b) Reconstructed results at night, poor semantic map result due to insufficient lighting.}
  \label{fig:waymo_longtail}
  \vspace{-1em}
\end{figure*}

We present some long-tail cases in challenging driving conditions. Fig.~\ref{fig:waymo_longtail} shows reconstructed results of WOD in rainy and night conditions. It is clear that the reconstructed photometric map is greatly affected due to lens soiling, but the semantic map can still be reconstructed well. In the case of poor lighting conditions, semantic segmentation is concerned, so the marking of road surface elements is unclear in the constructed semantic map.

\begin{figure*}[t!]
    \includegraphics[width=1.0\textwidth]{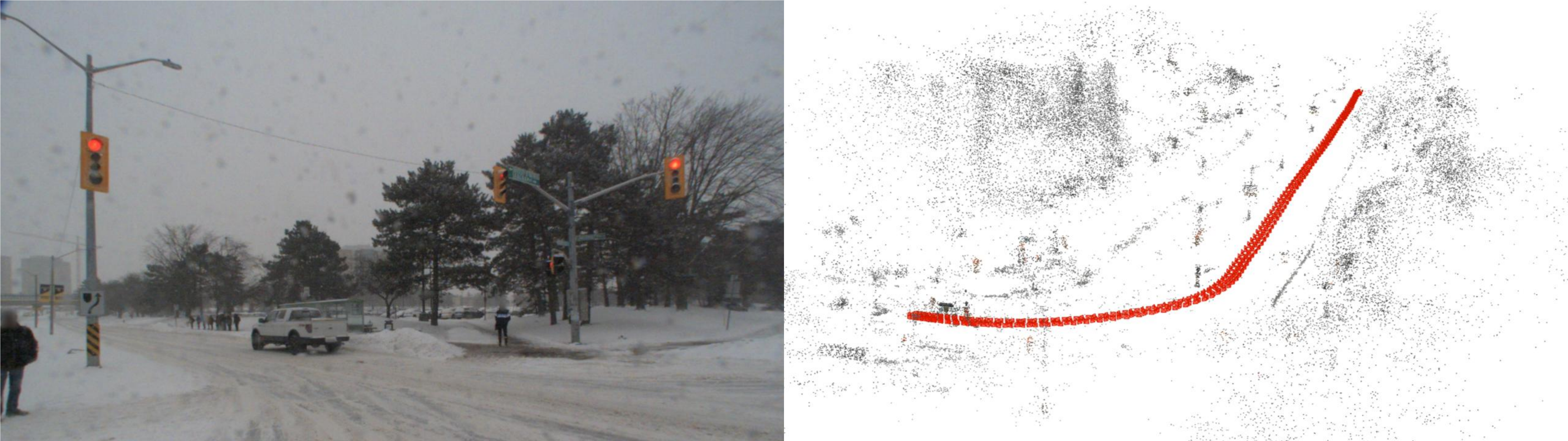}
  \caption{CADC reconstruction result (a challenging snow scenario) in the SfM stage.}
  \label{fig:cadc}
  \vspace{-1em}
\end{figure*}

To evaluate the effectiveness of our method for reconstruction in more extreme snowy scenes, we performed on the CADC dataset~\cite{pitropov2021canadian}. The snow-covered road surface poses a great challenge to the feature extraction task, leading to the failure of the road surface initialization. However, our method still achieves better results in the SfM stage (See Fig.~\ref{fig:cadc}).

\subsection{Dirty Details}
Some dirty details are proposed to improve the annotation quality: (1) To balance computational efficiency with sufficient visual coverage, we use consecutive video frames (called samples and sweeps) provided by nuScenes as input. Before the reconstructions, a keyframe selection step is conducted, where we set the maximum timestamp difference (40ms) for multi-view resynchronization. More dense keyframes are generated based on the relative change in ego-pose between frames (e.g., translation distance, rotation degree). (2) Some driving clips in nuScenes do not provide variation observations because the ego vehicle is stationary. Therefore, SfM reconstruction is not available. To solve this problem, we combine these stationary clips with their nearest long-driving clip. All frames of the merged clip are fed into CAMAv2 for reconstruction (without keyframe selection). Local map annotations of these stationary clips can be efficiently obtained using this method. (3) In the nuScenes dataset, the trunk is visible in the rear camera. We masked this part to avoid any impact on feature extraction and matching. (4) The parameter $\Delta\mathbf{z}^{t}$ can be empirically set to around $3m$ to ensure the accuracy of filtering matching pairs.

\section{Conclusion}
\label{sec:conclusion}
We present CAMAv2: a vision-centric approach for consistent and accurate static map element annotations. We investigate the critical factors for the online HD map construction algorithm and argue that the annotation quality in terms of reprojection accuracy and spatial-temporal consistency is vital for perception algorithm training. Based on this insight, we propose a new baseline for the BEV perception algorithm. Experiments on the nuScenes datasets show that our proposed methods not only generate high-quality annotations respecting accuracy and consistency but also improve the performance of perception models trained with our annotation. By making the CAMAv2 source codes and nuScenes-CAMAv2 annotation publicly available, we aim to catalyze advancements in intelligent driving and 4D labelling technologies within both academic and industrial communities.

\section*{Acknowledgments}
This work was supported in part by the Natural Science Foundation of the Jiangsu Higher Education Institutions of China (22KJB520008), and in part by the Research Fund of Horizon Robotics (H230666).

\bibliographystyle{IEEEtran}
\bibliography{ref}

\vspace{40 mm}
\begin{IEEEbiography}[{\includegraphics[width=1in,height=1.25in,clip,keepaspectratio]{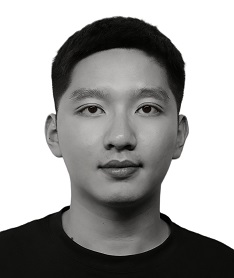}}]{Shiyuan Chen} 
is currently a postgraduate student at Soochow University. He earned his B.E. degree in Software Engineering from the Tiangong University in 2023. His primary research interests lie in 3D vision and deep learning.
\end{IEEEbiography}

\begin{IEEEbiography}[{\includegraphics[width=1in,height=1.25in,clip,keepaspectratio]{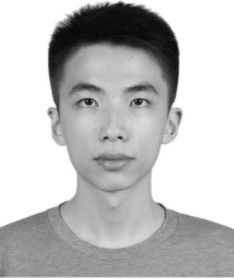}}]{Jiaxin Zhang}
is currently an Algorithm Engineer at Horizon Robotics in Beijing, China. He earned his B.S. degree in Applied Physics from the University of Science and Technology of China in 2018, followed by an M.S. degree in Electrical and Computer Engineering from Boston University in 2020. His primary research interests lie in SLAM, 3D vision, and deep learning.
\end{IEEEbiography}

\begin{IEEEbiography}[{\includegraphics[width=1in,height=1.25in,clip,keepaspectratio]{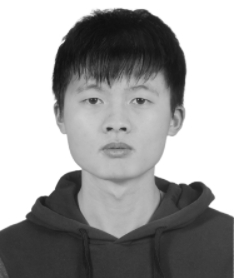}}]{Ruohong Mei}
is currently an Algorithm Engineer at Horizon Robotics in Beijing, China. He earned his B.S. Degree in Communication Engineering from Beijing University of Posts and Telecommunications in 2018, followed by an M.S. Degree in Information and Communication Engineering from Beijing University of Posts and Telecommunications in 2021. His primary research interests lie in 3D vision, and deep learning.
\end{IEEEbiography}

\begin{IEEEbiography}[{\includegraphics[width=1in,height=1.25in,clip,keepaspectratio]{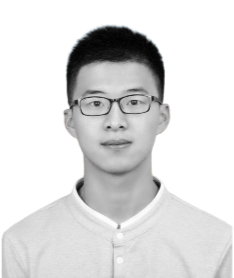}}]{Yingfeng Cai}
is currently an Algorithm Engineer at Horizon Robotics in Beijing, China. He earned his B.S. degree from Shanghai Maritime University, Shanghai, China, in 2020, followed by an M.S. degree in Computer Science and Technology from Tongji University, Shanghai, China, in 2023. His primary research interests lie in SLAM, 3D vision, and place recongnition.
\end{IEEEbiography}

\begin{IEEEbiography}[{\includegraphics[width=1in,height=1.25in,clip,keepaspectratio]{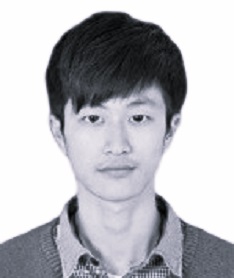}}]{Haoran Yin}
is currently an Algorithm Engineer at Horizon Robotics in Beijing, China. He earned his B.S. degree in Communication Engineering from Jilin University, Changchun, China, in 2019, and his M.S. degree in Computer Technology from the University of the Chinese Academy of Sciences, Beijing, China. His research interests include 3D vision, end-to-end autonomous driving, Vision Transformers, and AutoML.
\end{IEEEbiography}

\begin{IEEEbiography}[{\includegraphics[width=1in,height=1.25in,clip,keepaspectratio]{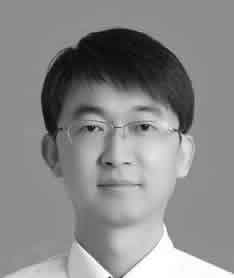}}]{Tao Chen}
received the B.Sc. degree in Mechanical Design, Manufacturing and Automation, M.Sc. degree in Mechatronic Engineering, and Ph.D. degree in Mechatronic Engineering from Harbin Institute of Technology, Harbin, China, in 2004, 2006, and 2010, respectively. He is a visiting scholar in National University of Singapore in 2018. He is currently an professor at School of Future Science and Engineering, Soochow University, Suzhou, China. His main research interests include MEMS, sensors, and actuators.
\end{IEEEbiography}

\begin{IEEEbiography}[{\includegraphics[width=1in,height=1.25in,clip,keepaspectratio]{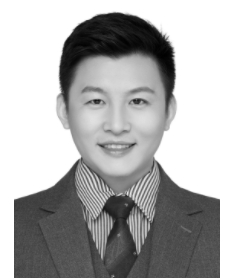}}]{Wei Sui}
is a senior engineer at Horizon Robotics, leading the 3D Vision Team, providing mapping, localization, calibration, and 4D labeling solutions. His research interests include SFM, SLAM, Nerf, 3D Perception, etc. Dr. Wei received his B.Eng and Ph.D. degrees from Beinghang University and NLPR (CASIA), Beijing, China, in 2011 and 2016 respectively. He led the computer vision team and successfully developed the 4D Labeling System and BEV perception for Super Drive on Journey 5. Dr. Wei Sui has published one research monograph and more than ten peer-reviewed papers in journals and conference proceedings, including elites like TIP, TVCG, ICRA, CVPR, etc. Dr. Wei received over 40 Chinese and 5 US patents.
\end{IEEEbiography}

\begin{IEEEbiography}[{\includegraphics[width=1in,height=1.25in,clip,keepaspectratio]{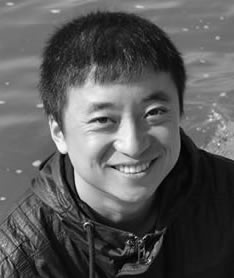}}]{Cong Yang}
is an Associate Professor at Soochow University since 2022. Before that, he was a Postdoc researcher at the MAGRIT team in INRIA (France). Later, he worked scientifically and led the computer vision and machine learning teams in Clobotics and Horizon Robotics. His main research interests are computer vision, pattern recognition, and their interdisciplinary applications. Cong earned his Ph.D. degree in computer vision and pattern recognition from the University of Siegen (Germany) in 2016.
\end{IEEEbiography}

\vfill

\end{document}